\documentclass[letterpaper, 10 pt, conference]{ieeeconf}  

\IEEEoverridecommandlockouts                              

\title{\LARGE \bf
Guiding Reinforcement Learning with Incomplete System Dynamics
}

\author{Shuyuan Wang$^{1}$, Jingliang Duan$^{4}$, Nathan P. Lawrence$^{2}$, Philip D. Loewen$^{2}$\\Michael G. Forbes$^{3}$, R. Bhushan Gopaluni$^{1}$, Lixian Zhang$^{5}$ \textit{Fellow IEEE}
\thanks{*This work was supported by NSERC and Honeywell Connected Plant.}
\thanks{Video available at \texttt{https://youtu.be/xGNNiuYJh98}.}
\thanks{$^{1}$Department of Chemical and Biological Engineering, University of British Columbia, Vancouver, Canada, {\tt\small wshuyuan@student.ubc.ca, bhushan.gopaluni@ubc.ca}}%
\thanks{$^{2}$Department of Mathematics, University of British Columbia, Vancouver, Canada, {\tt\small lawrence@math.ubc.ca, loew@math.ubc.ca}}%
\thanks{$^{3}$Honeywell Process Solutions, North Vancouver, Canada,
{\tt\small michael.forbes@honeywell.com}}
\thanks{$^{4}$School of Mechanical Engineering, University of Science and Technology Beijing, Beijing, China, {\tt\small duanjl@ustb.edu.cn}}
\thanks{$^{5}$Department of Astronautics, Harbin Institute of Technology, Harbin, China,  {\tt\small lixianzhang@hit.edu.cn}}
}
\usepackage{amsmath}
\usepackage{amssymb}
\usepackage{color}
\usepackage{algorithm}
\usepackage{algorithmic}
\usepackage{caption}
\usepackage{subcaption}
\usepackage{graphicx}
\usepackage{tabularx}
\usepackage{booktabs}
\usepackage{stfloats}
\usepackage[capitalize]{cleveref}
\usepackage{url}
\usepackage{boldline}
\makeatletter
\newcommand{\removelatexerror}{\let\@latex@error\@gobble}
\newtheorem{prop}{Proposition}
\makeatother

\usepackage{mathtools}

\usepackage{xcolor,color}
\definecolor{blu}{rgb}{0,0,1}

\definecolor{gre}{rgb}{0,.5,0}

\definecolor{red}{rgb}{1,0,0}

\usepackage{flushend}

\begin{document}

\maketitle
\thispagestyle{empty}
\pagestyle{empty}

\begin{abstract}

Model-free reinforcement learning (RL) is inherently a reactive method, operating under the assumption that it starts with no prior knowledge of the system and entirely depends on trial-and-error for learning. This approach faces several challenges, such as poor sample efficiency, generalization, and the need for well-designed reward functions to guide learning effectively.
On the other hand, controllers based on complete system dynamics do not require data.
This paper addresses the intermediate situation where 
there is not enough model information for complete controller design, but there is enough to suggest
that a model-free approach is not the best approach either.
By carefully decoupling known and unknown information about the system dynamics, we obtain an embedded controller guided by our partial model and thus improve the learning efficiency of an RL-enhanced approach. A modular design allows us to deploy mainstream RL algorithms to refine the policy.
Simulation results show that our method significantly improves sample efficiency compared with standard RL methods on continuous control tasks, and also offers enhanced performance over traditional control approaches.
Experiments on a real ground vehicle also validate the performance of our method, including generalization and robustness.

\end{abstract}

\section{Introduction}
Human learning, such as riding a bicycle, is often accelerated by leveraging accumulated prior knowledge about the dynamics of the world. This ability allows adults to learn new tasks through a relatively small number of trials, emphasizing the value of prior knowledge in efficient learning \cite{lake2017building}. Humans' dynamic process of trial and error bears a striking resemblance to reinforcement learning (RL) \cite{sutton2018reinforcement}. However, despite these similarities, RL agents typically learn from scratch, which necessitates a large number of interactions with the environment. This limitation blocks RL’s application toward real-world continuous control tasks.

\begin{figure}[thpb]
    \centering
    \includegraphics[width=0.53\textwidth]{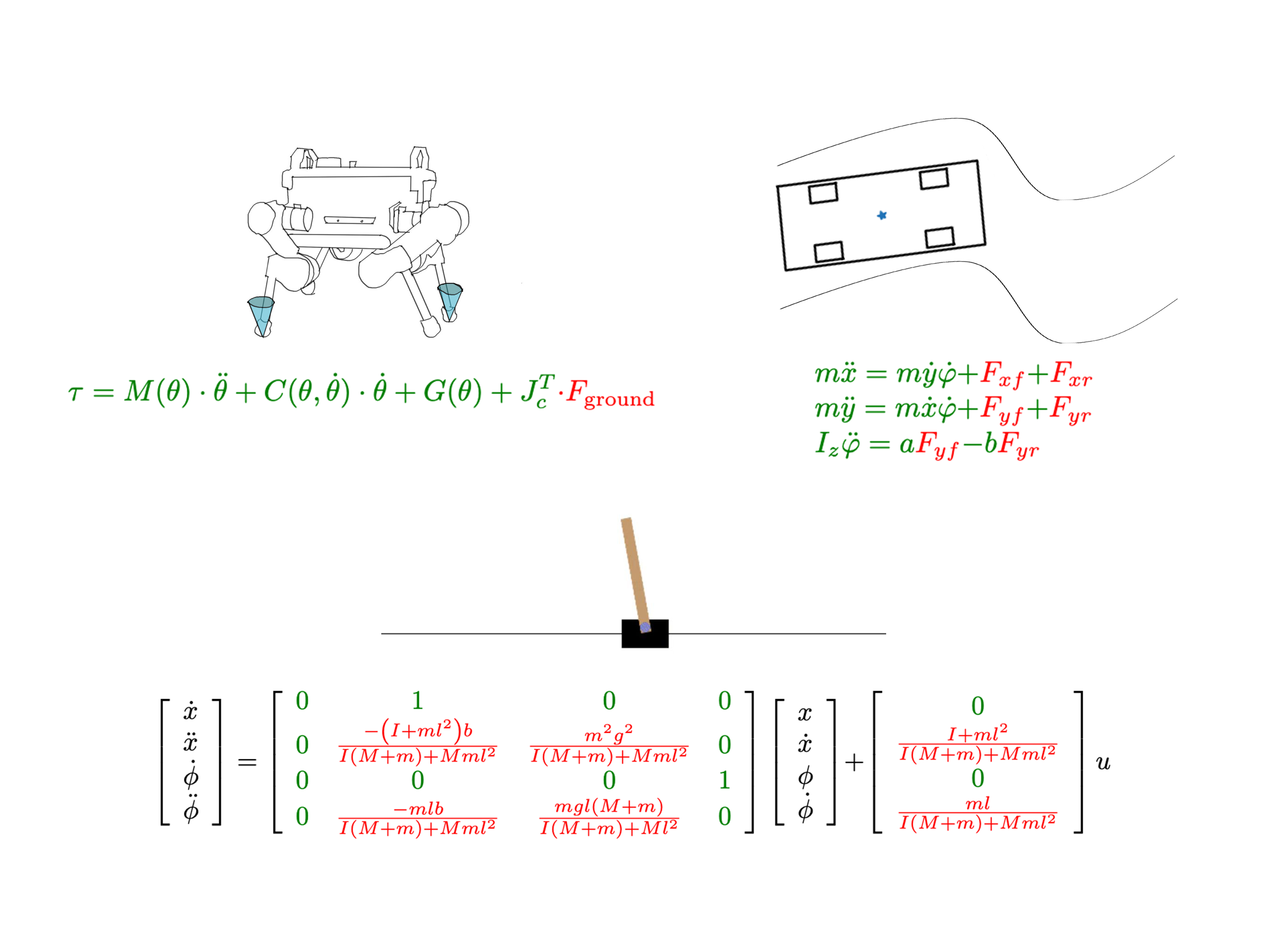}
    \caption{Dynamics model and partial model knowledge for different tasks: quadruped robot; self-driving vehicle; inverted pendulum. Green elements represent known parameters and structure; Red elements represent unknown parameters.}
    \label{fig:models}
\end{figure}

The current work seeks to narrow this gap by enhancing RL algorithms with partial system knowledge. In particular, we focus on incorporating prior knowledge related to model structure and some parameters into RL algorithms. To illustrate, consider the inverted pendulum system in \cref{fig:models}, a classic control problem. In this system, certain parameters related to the mechanical properties of the system (like length and inertia) are unknown, while the model structure and other parameters are known exactly, such as the 0 and 1 elements in the matrices. Most RL methods do not take such information into account.

In this work, we introduce a control module into RL policies that harnesses partial model knowledge, thereby aiming to improve both the sample efficiency and generalizability of RL algorithms. Our proposed method stands out by effectively decoupling known and unknown information. This results in a more efficient use of model information and an expected improvement in sample efficiency. Our approach integrates partial model knowledge into RL while preserving the core RL structure. Thus, it can be readily implemented with mainstream RL methods. Empirical results from simulations and real-world robot experiments show that our approach outperforms several prominent RL algorithms.

We highlight the following contributions:
\begin{enumerate}
 \item We introduce a novel framework that brings partial model knowledge into RL in a decoupled way, bridging the RL and control frameworks without disrupting the RL structure.
    \item Our method enhances RL's sample efficiency, consistency, and generalization while maintaining its capacity for optimality.
\end{enumerate}

\section{Related works}

\subsection{Reinforcement learning and adaptive control} 
RL algorithms have proven to be highly effective in tackling complex decision-making and control tasks through trial-and-error.
However, such data collection usually entails high sample complexity, which limits the feasibility of RL algorithms when applied in real-world applications \cite{cobbe2019quantifying,yao2021sample}. 

Various enhancements have been developed to bolster RL performance, such as double $Q$-learning \cite{van2016deep} for mitigating overestimation, proximal policy optimization \cite{schulman2017proximal} for safe updates, and soft iterations \cite{haarnoja2018soft} for exploration. 
Nonetheless, RL's nature of learning from scratch remains unchanged, impeding faster learning and generalization. Our approach breaks the policy network `black box' by integrating a control module that utilizes partial model knowledge, thus enhancing learning without altering the existing framework. This strategy maintains existing frameworks' strengths and also capitalizes on the benefits of control and partial models for more efficient learning.

Our method is distinct from model-based RL \cite{deisenroth2011pilco, heess2015learning}, which constructs a model from scratch to generate synthetic data or estimate cost gradients relative to policy parameters. Our approach diverges by leveraging known structures and parameters a priori, rather than building from zero. We aim to incorporate this knowledge into a model-free framework, merging model-free RL's flexibility with control theory's precision. Notably, model-based RL methods such as ME-TRPO \cite{kurutach2018modelensemble}, SLBO \cite{luo2018algorithmic}, and MBPO \cite{pmlr-v87-clavera18a} still utilize model-free RL for policy optimization after model creation, indicating that improvements in model-free techniques could benefit model-based strategies. Thus, our method complements rather than conflicts with existing paradigms.

In the context of adaptive control, \cite{modares2014integral,VRABIE2009477} explored policy iteration with partial knowledge. These works assume the control matrix $B$ is fully known and the state transition matrix $A$ is fully unknown for linear systems $x_{k+1}=Ax_k+Bu_k$. Although the titles include `partial model', these assumptions narrow the scope of partial knowledge to a specific subset, limiting its breadth in the domain of partial models.


\subsection{End-to-end learning} 
In the context of control and planning, end-to-end learning optimizes any parameters (such as policy parameters, and model parameters) directly from the overall performance metric. As a milestone, \cite{vin} shows that embedding the planning module into an RL policy network can augment the policy's generalizability, leading to acceptable performance even in unexplored domains. However, \cite{vin} assumes discrete action and state spaces. 
Differentiable model predictive control (MPC) \cite{differentiableMPC,amos2017optnet,east2020infinite} provides insights to incorporate a continuous-action control module into a network. However, these approaches assume a linear model structure and quadratic cost. These issues make the approaches fall short in handling model biases and flexible rewards, thereby limiting their application to RL tasks. Our method is built upon these methods but bypasses these limitations by introducing a compensation structure, enabling control to dance better with RL.

\section{Preliminary: Differentiable MPC}




In order to incorporate partial knowledge-based control within the reinforcement learning framework, we introduce a differentiable control module as an integrated layer within the RL policy network. At the core of this module lies a linear MPC strategy, formulated as follows:
\begin{equation}
x(t+dt)-x_d(t+dt)=A(x(t)-x_d(t))+B(u(t)-u_d(t)),
\end{equation}
where $x(t)$ represents the system state at time $t$, $u(t)$ denotes the control input, and $x_d(t)$ and $u_d(t)$ correspond to the desired state and control input, respectively. When $x_d$ and $u_d$ are set to zero, the problem reduces to the well-known Linear Quadratic Regulator (LQR). The objective is to minimize a quadratic stage cost: 
\begin{equation}
\begin{aligned}
    J = &\sum_{t=0}^{\infty} (x(t)-x_d(t))^\top Q(x(t)-x_d(t))\\
    &+(u(t)-u_d(t))^\top R(u(t)-u_d(t))
\end{aligned}
\end{equation}
with weight matrices $Q\geq0$ and $R>0$ specifying the importance of state and control input deviations in the cost function. In our framework, the weights can be aligned with the rewards from the RL environment or predefined based on common practices. As shown in the experiment section, our proposed framework can adapt well to various kinds of objective functions from the environment.


The action $u$ that optimizes the accumulated cost is obtained by solving the Discrete Algebraic Riccati Equation (DARE)
\begin{equation}
	P=Q+A^\top PA-A^\top PB(R+B^\top PB)^{-1}B^\top PA,
 \label{DARE}
\end{equation}
and the optimal controller is given by
\begin{equation}
    u=-(R+B^\top PB)^{-1}B^\top PA (x-x_d)+u_d.
\end{equation}

To integrate linear MPC as a differentiable layer within the RL framework, it 
we need to know the sensitivity of
the solution matrix $P$ with respect to the dynamics $A$ and $B$.
As shown in \cite{east2020infinite}, this can be achieved by differentiating through (\ref{DARE}) and solving for a closed-form solution. 
Here is the result.
\begin{prop}[\cite{east2020infinite}]
\label{SensitivityProposition}
Let $P$ be the stabilizing solution of DARE, and assume that $Z_1^{-1}$
and $(R + B^\top P B)^{-1}$
exists. Then the Jacobians of the implicit function defined by DARE are given by
\begin{equation}
\begin{aligned}
\frac{\partial \operatorname{vec} P}{\partial \operatorname{vec} A}=Z_1^{-1} Z_2, \quad \frac{\partial \mathrm{vec} P}{\partial \mathrm{vec} B}=Z_1^{-1} Z_3, 
\end{aligned}
\end{equation}
where $Z_1,Z_2, Z_3$ are defined by
\begin{equation}
\begin{aligned}
& Z_1= \mathbf{I}_{n^2}-\left(A^{\top} \otimes A^{\top}\right)\left[\mathbf{I}_{n^2}-\left(P B M_2 B^{\top} \otimes \mathbf{I}_n\right)\right. \\
&\left.\quad +(P B \otimes P B)\left(M_2 \otimes M_2\right)\left(B^{\top} \otimes B^{\top}\right)\right. \\
& \left.\quad-\left(\mathbf{I}_n \otimes P B M_2 B^{\top}\right)\right]\\
& Z_2=\left(\mathbf{V}_{n, n}+\mathbf{I}_{n^2}\right)\left(\mathbf{I}_n \otimes A^{\top} M_1\right) \\
& Z_3=\left(A^{\top} \otimes A^{\top}\right)\left[(P B \otimes P B)\left(M_2 \otimes M_2\right) \right.\\ 
&\left.\quad\left(\mathbf{I}_{m^2}+\mathbf{V}_{m, m}\right)\left(\mathbf{I}_m \otimes B^{\top} P\right)\right. 
\end{aligned}
\label{Z123}
\end{equation}
and $M_1, M_2, M_3$ are defined by
\begin{equation*}
M_1:=P-P B M_2 B^{\top} P, M_2:=M_3^{-1}, M_3:=R+B^{\top} P B.
\end{equation*}
\end{prop}


\section{Main Method}
\subsection{Partial Knowledge Representation}
We consider the dynamical system model in continuous time, given by:
\begin{equation}
\dot{x}=f(x,u).
\end{equation}
We adopt a continuous-time model, as it often captures the physical properties of the system and thus contains inherent prior information. Note that this does not limit our method's ability to adapt to discrete-time systems. We will elaborate on this point later in this section.

In order to formulate known information and unknown information, we decompose the model as follows:
\begin{equation}
f(x,u)=f_{app}(f_1(x,u),f_2(x,u))+f_{bias}(x,u).
\label{partial model}
\end{equation}
Here, $f_{app}$ represents the approximated model with known structure, while $f_{bias}(x,u)$ accounts for the bias of the approximate model. Within $f_{app}(x,u)$, we have both known information $f_1(x,u)$ and unknown information $f_2(x,u)$, and we also know how $f_1$ and $f_2$ combine to form $f_{app}$. In essence, $f_2(x,u)$ and $f_{bias}(x,u)$ make up the unknown part of the overall model. Taking the inverted pendulum example from Fig.\ref{fig:models}, \(f_{app}\) represents the linearized model of the system. \(f_{bias}\) captures the error introduced by this linearization, which is not explicitly represented or learned within our framework. \(f_1\) corresponds to the known effects of the input, represented by zeros and ones in the matrix, while \(f_2\) deals with the impact of unknown elements in the matrix on the input. The unknown parameters within \(f_2\) are denoted by \(\psi\), making \(f_{app}\) rewritten as \(f_{app}(\psi)\). Our framework focuses on \(f_{app}(\psi)\), leveraging the capabilities of RL to learn control strategies that effectively address the uncertainties and biases represented by \(f_2\) and \(f_{bias}\).

\subsection{Framework}



\begin{figure*}[thpb!]
    \centering
    \includegraphics[width=0.85\textwidth]{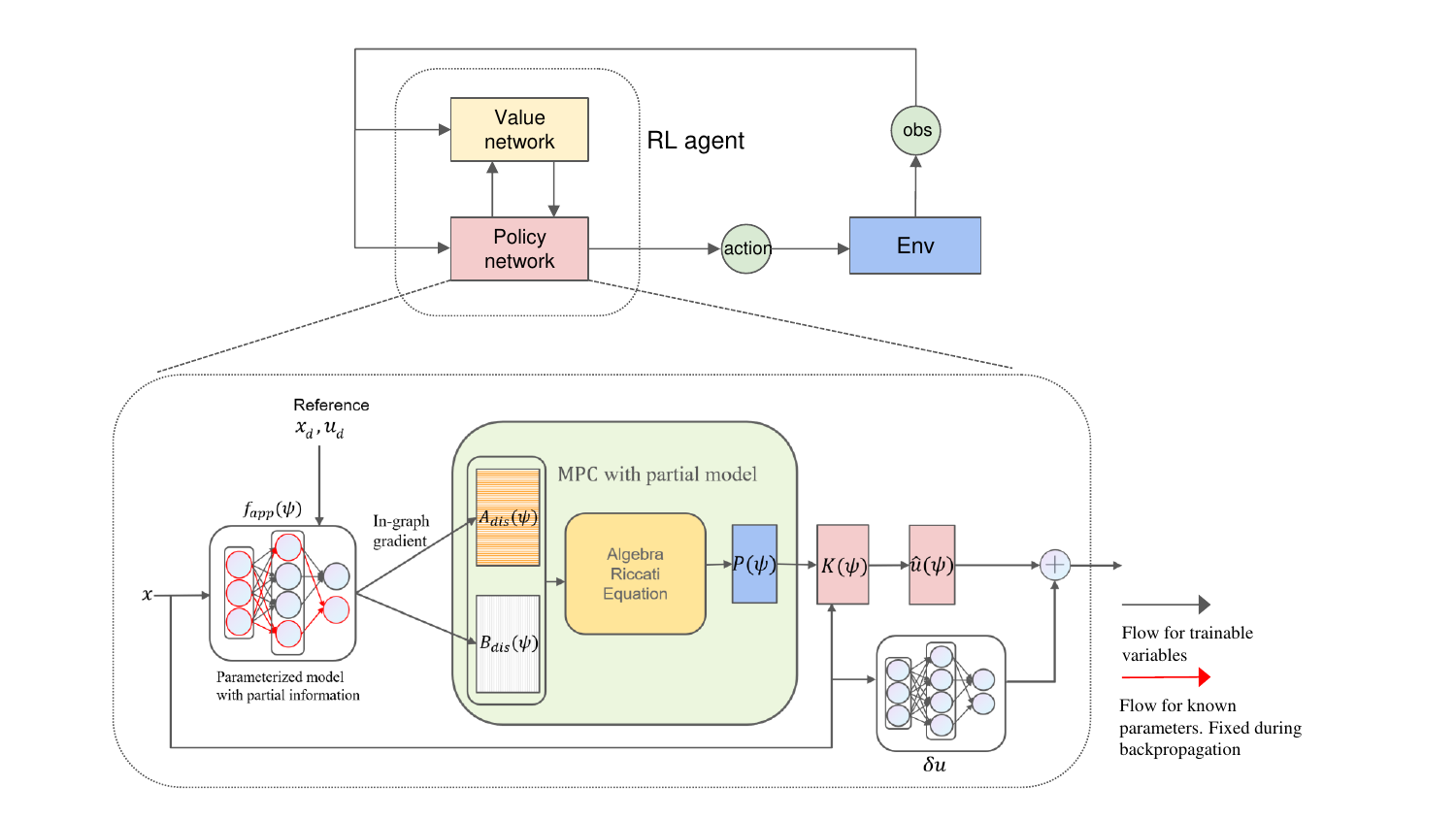}
    \caption{Schematic diagram for our policy network with partial knowledge control module inside.}
    \label{policy_net}
\end{figure*}

We construct a novel policy network that incorporates a differential control module. This module enables planning based on partial knowledge and facilitates training within an RL framework. To address the action errors stemming from the model's bias and uncertainties associated with unknown parameters, an additive, separate neural network is introduced for compensation, thereby enhancing performance. Together, the policy with partial system knowledge can be deployed in a standard model-free RL pipeline. 
\cref{policy_net} illustrates our framework.

In the module, \(f_{app}\) is built as a differentiable computational graph, with \(\psi\) representing its unknown but trainable parameters. The module first linearizes the nonlinear model around the reference trajectory or setpoint, $x_d$ and $u_d$. This is done by taking the \textbf{\textit{in-graph gradient}} with respect to $x_d$ and $u_d$. The output is a linear model with coefficient matrices given by
\begin{equation}
    A(\psi)=\frac{\partial f_{app}}{\partial x}\Big|_{x_d}, B(\psi)=\frac{\partial f_{app}}{\partial u}\Big|_{u_d}.
    \label{linearaztion}
\end{equation}
Given step size $\tau$, we discretize the continuous model with the Euler method:
\begin{equation}
    A_{\rm dis}(\psi)=\tau A(\psi)+I, B_{\rm dis}(\psi)=\tau B(\psi)
\end{equation}
where $I$ is the identity matrix with the same dimension as $A(\psi)$. If the system (\ref{partial model}) originally operates in discrete time, the linearization directly applies to the model with partial knowledge, bypassing the need for discretization
\begin{equation}
    A_{\rm dis}(\psi)=\frac{\partial f_{app}}{\partial x}\Big|_{x_d}, B_{\rm dis}(\psi)=\frac{\partial f_{app}}{\partial u}\Big|_{u_d}.
\end{equation} This direct linearization simplifies the adaptation of our method to inherently discrete systems.

With $A_{\rm dis}(\psi)$ and $B_{\rm dis}(\psi)$, the parameterized value matrix $P(\psi)$ can be obtained by solving the DARE (\ref{DARE}). The gradient of $P$ with respect to $\psi$ will be introduced in the section \ref{sec:Training}. With $P(\psi)$, the module calculates a sub-optimal baseline action by  
\begin{equation}
     \hat{u}(\psi)=-K(\psi)(x-x_d)+u_d,
\end{equation}
where $K(\psi)$ is given by
\begin{equation}
    (R+B_{\rm dis}(\psi)^\top P(\psi)B_{\rm dis}(\psi))^{-1}B_{\rm dis}(\psi)^\top P(\psi)A_{\rm dis}(\psi),  
\end{equation}
which follows exactly the scheme of linear MPC.

In contrast to existing differentiable LQR and MPC methods \cite{east2020infinite,differentiableMPC}, our framework can mitigate the model bias. This is achieved by introducing the corrective structure
\begin{equation}
    u=\hat{u}+\delta u.
\end{equation}
Here $\delta u$ is generated from a trainable neural network, with input directly as the state. The $\delta u$ is used to address the suboptimality introduced by the linear control with unknown parameters, and the model bias term $f_{bias}(x,u)$ from the original nonlinear model and the linearization error from (\ref{linearaztion}). We hypothesize that this approach will enjoy a lower learning burden and hence faster learning speed. For deterministic policy in DDPG and TD3, $\delta u$ is output from a network directly. For stochastic policy in SAC and PPO, the output of the network is the mean and standard deviations of a Gaussian distribution. We admit that the initial value and magnitude of $\delta u$ may impact the training outcomes. At this early stage, we have not yet explored theoretical guidance for these parameters. In our experiments, we set the initial value of $\delta u$ to $0$ in order to emphasize the dominant role of $\hat{u}$.




\subsection{Training}
\label{sec:Training}

All the operations in the proposed framework are differentiable, and consequently, the whole process can be trained through backpropagation. 
To separate the contribution from known parameters and unknown parameters, \textit{\textbf{our framework keeps the known parameters fixed and only backpropagates the gradient through the unknown parameters}}. In this way, the contributions from different information are decoupled. Therefore, RL only learns the policy with respect to unknown information.

For the backpropagation within MPC, we use the chain rule to calculate the gradient:
\begin{equation}
    \frac{\partial P}{\partial \psi}=
    \frac{\partial P}{\partial A}\frac{\partial A}{\partial \psi}+
    \frac{\partial P}{\partial B}\frac{\partial B}{\partial \psi}.
\end{equation}

By keeping the known model parameters constant, the gradients of $P,A,B$ w.r.t. parameters outside of $\psi$ are eliminated. The $\frac{\partial A}{\partial \psi}$ and $\frac{\partial B}{\partial \psi}$ can be obtained using Automatic Differentiation (AD) tools provided in standard packages such as PyTorch \cite{paszke2019pytorch} and TensorFlow \cite{tensorflow2015-whitepaper}. Combined with the analytical solutions of $\frac{\partial P}{\partial A}$ and $\frac{\partial P}{\partial B}$ from Prop.~\ref{SensitivityProposition}, the gradient of $P$ with respect to unknown parameters is calculated.

\section{Experiments}

\subsection{Simulation results}

We evaluate our method using tasks from OpenAI gym \cite{brockman2016openai} and the MuJoCo physics engine \cite{todorov2012mujoco}: CartPole, and Inverted Double Pendulum, which are widely recognized benchmarks in control and RL. The goal of the tasks can be summarized as achieving an expected configuration.

\begin{table*}[t]
\centering
\begin{tabular}{cc}
    CartPole & Inverted Double Pendulum \\
    \includegraphics[width=0.20\linewidth]{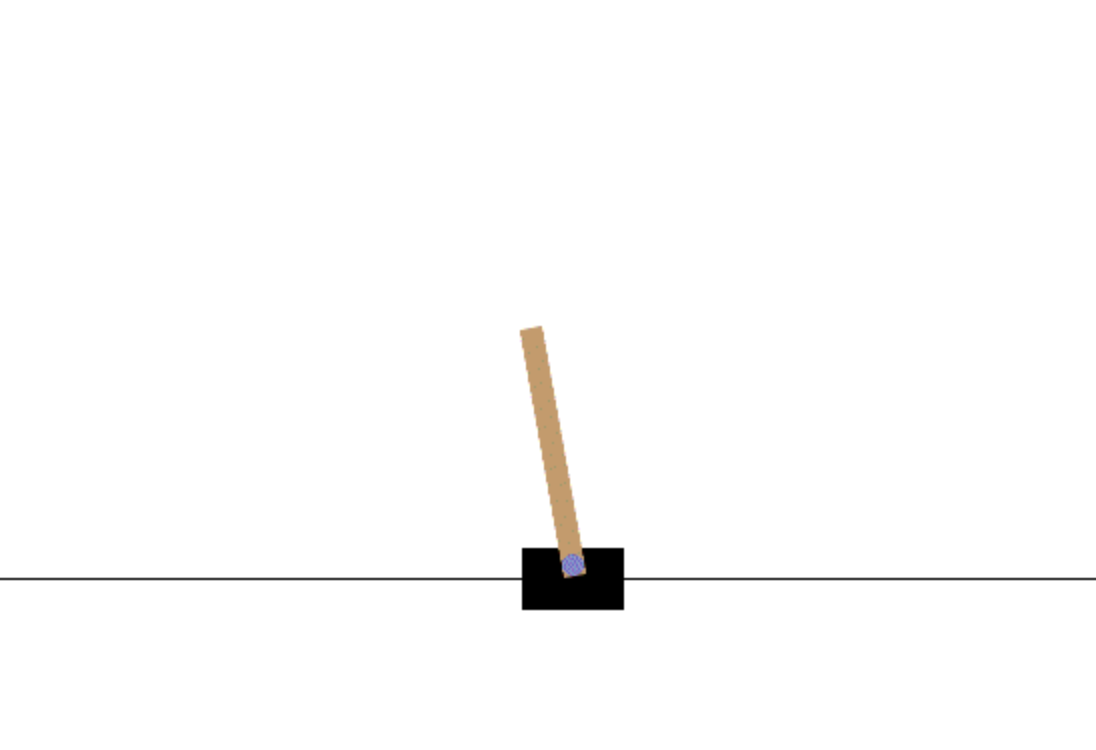} & 
    \includegraphics[width=0.15\linewidth]{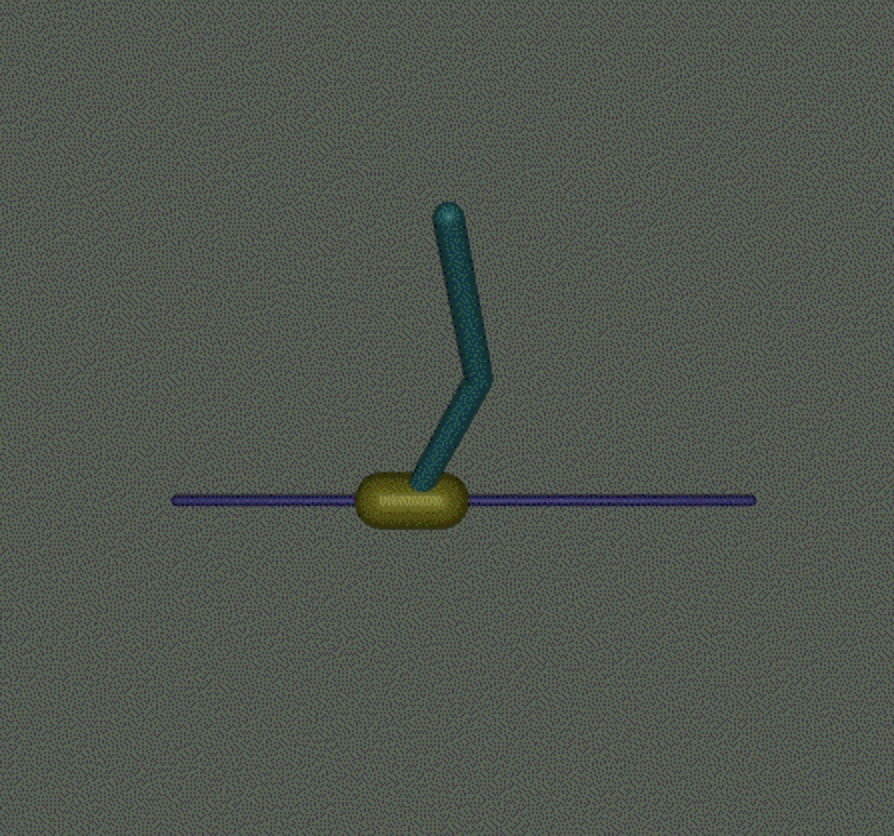} \\
    \hline
    $\begin{aligned}
    \ddot{\theta} &= \frac{g \sin(\theta) - \cos(\theta) \left( \frac{u - m_p l \dot{\theta}^2 \sin(\theta)}{m_p + m_c} \right)}{\frac{1}{10} \left( \frac{4}{3} - \frac{m_p \cos(\theta)^2}{m_p + m_c} \right)} \\
    \ddot{x} &= \frac{u - m_p l \dot{\theta}^2 \sin(\theta)}{m_p + m_c} - \frac{m_p l  \ddot{\theta} \cos(\theta)}{m_p + m_c}
    \end{aligned}$ &
    $
    \begin{array}{r}
    \left(m_0+m_1+m_2\right) \ddot{x}+\left(0.5m_1L_1+m_2 L_1\right) \cos \theta_1 \ddot{\theta}_1+0.5m_2L_2 \cos \theta_2 \ddot{\theta}_2 \\
    -\left(0.5m_1L_1+m_2 L_1\right) \sin \theta_1 \dot{\theta}_1^2-0.5m_2L_2 \sin \theta_2 \dot{\theta}_2^2=u \\
    \left(0.25m_1 L_1^2+m_2 L_1^2+m_1L^2_1/12\right) \ddot{\theta}_1+\left(0.5m_1 L_1+m_2 L_1\right) \cos \theta_1 \ddot{\theta}_0 \\
    +0.5m_2 L_1 L_2 \cos \left(\theta_1-\theta_2\right) \ddot{\theta}_2+0.5m_2 L_1L_2 \sin \left(\theta_1-\theta_2\right) \dot{\theta}_2^2 \\
    -g\left(0.5m_1 L_1+m_2 L_1\right) \sin \theta_1=0 \\
    0.5m_2 L_2 \cos \theta_2 \ddot{\theta}_0+0.5 m_2 L_1 L_2 \cos \left(\theta_1-\theta_2\right) \ddot{\theta}_1+\left(0.25m_2 L_2^2+m_2L^2_2/12\right) \ddot{\theta}_2 \\
    -0.5m_2 L_1 L_2 \sin \left(\theta_1-\theta_2\right) \dot{\theta}_1^2-0.5m_2 g L_2 \sin \theta_2=0
    \end{array}
    $
\end{tabular}
\caption{ Visualizations and dynamics models of CartPole and Inverted Double Pendulum (IDP): Displacement \(x\) and angle \(\theta\) for the cart with cart mass \(m_c\) and pole mass \(m_p\); Displacement \(x\), angles \(\theta_1\) and \(\theta_2\), and masses \(m_0\) (the cart), \(m_1\) (the pole below), and \(m_2\) (the pole above) for Inverted Double Pendulum. All mechanical properties, such as mass and length, are unknown parameters. The models are derived from Lagrangian mechanics.}
\label{task_model}
\end{table*}


We demonstrate the following features of our method:
\begin{enumerate}
    \item Faster learning speed than its corresponding base RL method.
    \item Performance improvement over LQR.
    \item Adaptation to different reward functions.
\end{enumerate}

To better showcase the strength of our method, we have modified the original Gym environments. For CartPole, we retained the original environment's alive bonus of 1 but changed the action space from discrete to continuous. For the Inverted Double Pendulum (IDP), we removed the environment's original alive bonus of 10, relying solely on the quadratic cost of configuration and velocity to motivate the RL agent. 
Specifically, the reward is calculated as the negative sum of the distance penalty,  $x^2 + 2(y - 1.2)^2$,  where  $x$  and  $y$  are the coordinates of the pendulum tip, and the velocity penalty,  $10v_1^2 + 20v_2^2$,  where  $v_1$ and  $v_2$ are the angular velocities of the two pendulum joints. This encourages minimizing deviations from the upright position and penalizes excessive joint velocities.
Experiments demonstrate that traditional RL methods struggle to learn from quadratic costs effectively.

For our baseline algorithms, we selected SAC \cite{haarnoja2018soft}, TD3 \cite{fujimoto2018addressing}, and Policy Gradient (PG) \cite{sutton1999policy}. When our method is applied to these baselines, we denote them as PK-SAC, PK-TD3, and PK-PG, respectively. (PK stands for Partial Knowledge.)

For our partial knowledge-based method, we represent partial knowledge using the model structure as shown in Table \ref{task_model}, where all mechanical properties, such as mass and length, are treated as unknown parameters. Specifically, for the CartPole task, to demonstrate the limits of our approach, we intentionally initialize the model's unknown parameters with values that render the system unstable under a classic LQR controller. For the IDP task, we broaden the range of initial states in the environment to include configurations that are difficult for LQR to control effectively.
\begin{figure*}[tbp!]
    \centering
    \begin{subfigure}[b]{2.75in}
        \centering
        \includegraphics[width=\textwidth]{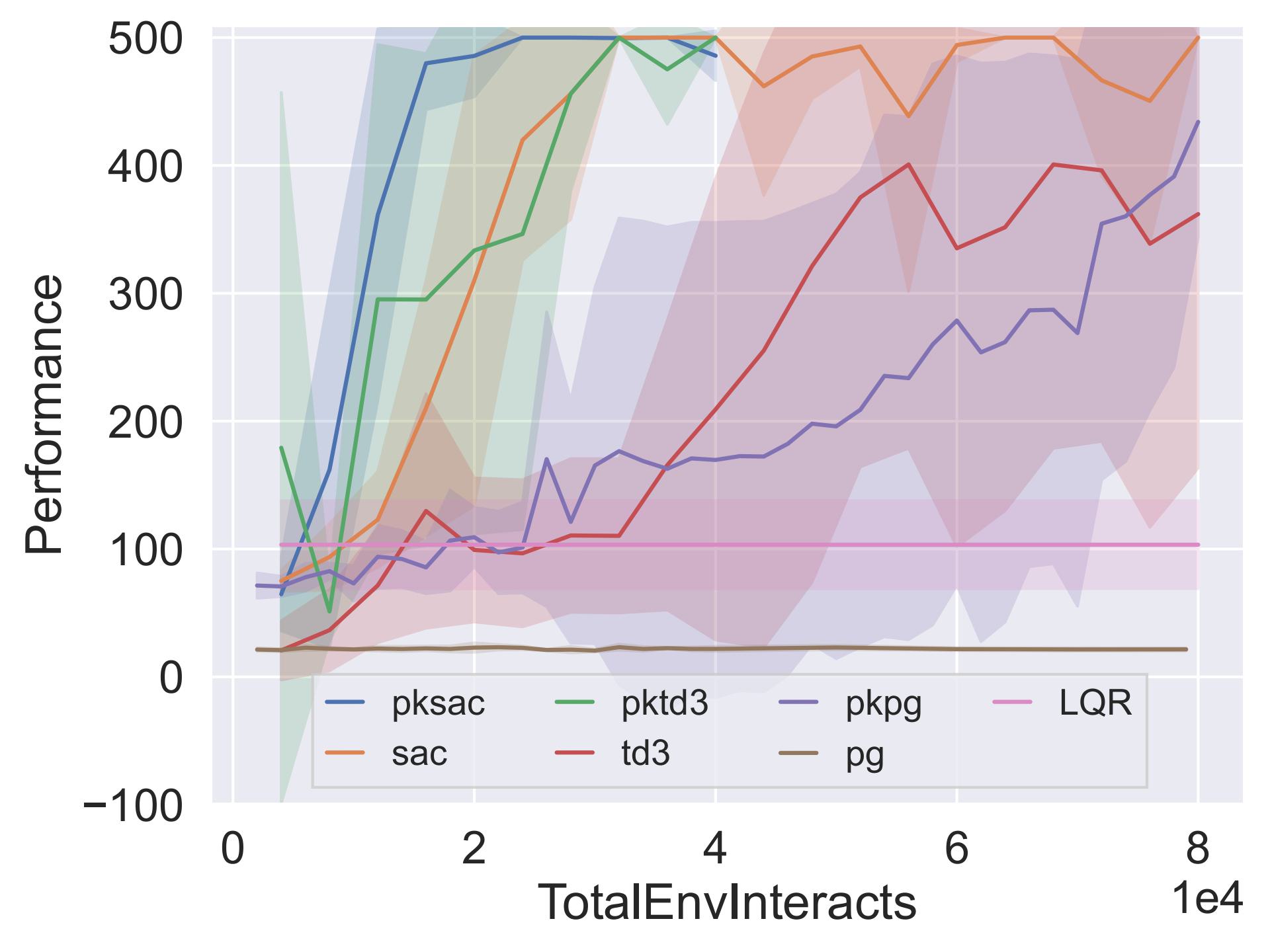}
        \caption{CartPole with only alive bonus.}
        \label{cartpole}
    \end{subfigure}
    \begin{subfigure}[b]{2.75in}
        \centering
        \includegraphics[width=\textwidth]{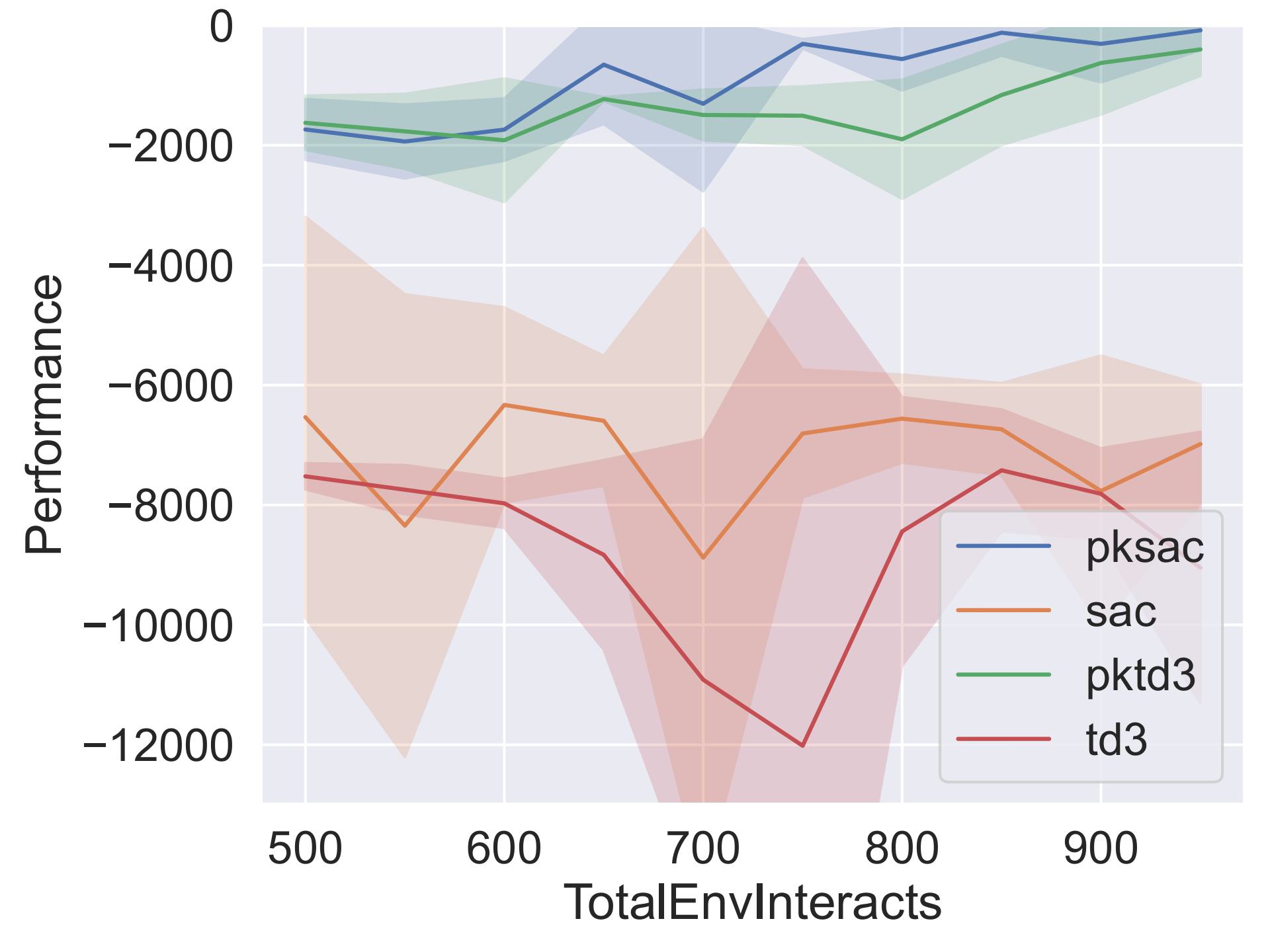}
        \caption{IDP with only quadratic cost.}
        \label{idp}
    \end{subfigure}
    \caption{Training curves on the control benchmarks. Solid lines show the mean; shaded regions show the standard deviations over five runs.}
    \label{curves}
    \vspace{-15pt}
\end{figure*}
The baseline methods are adopted directly from OpenAI Spinning Up (\url{https://github.com/openai/spinningup}), a renowned library recognized for its implementations of RL algorithms that consistently achieve state-of-the-art performance as reported in the literature.

We conducted five separate tests for each algorithm, using
distinct random seeds across all algorithms. All the experiments are conducted on a platform with AMD 3700X, 8 core, 3.6GHZ, and 16G memory.

\subsubsection{Enhanced Learning Speed}

\Cref{curves} compares all algorithms in terms of the performance. This comparison demonstrates that our PK-based method can achieve superior performance with significantly less data than the baselines, demonstrating the enhanced sample efficiency of our approach.
\begin{table}[tbp]
    \centering
    \caption{Cost comparison between our method and LQR on IDP.}
    \resizebox{0.38\textwidth}{!}{
    \begin{tabular}{l|ccc}
        \toprule 
        Initial velocities & PKSAC & PKTD3 & LQR \\
        \midrule 
        $[-0.13,-0.13,0.17]$ & $-1012.54$ & $\textbf{-731.33}$ &$-10444.72$  \\
        $[-0.11,-0.13,0.17]$ & $-793.89$ & $\textbf{-720.17}$ & $-11635.68$ \\
        $[-0.13,-0.13,0.18]$ & $-864.21$ & $\textbf{-709.86}$ & $-10973.45$\\
        \bottomrule
    \end{tabular}}
    \label{tab:performance_comparison}
    \vspace{-15pt}
\end{table}
Notably, our framework improves the performance of algorithms that previously failed to learn in more complex settings. We applied Policy Gradient (PG) to our CartPole task, which is traditionally used as a fundamental benchmark within discrete action spaces. As shown in \cref{cartpole}, PG was unsuccessful when adapted to continuous action spaces, demonstrating a significant gap in its applicability. However, by integrating our partial model framework into PG, thereby creating PKPG, we significantly enhance its learning capability and stimulate it to control the system successfully.

Similarly, in the IDP experiments that relied solely on quadratic cost, baseline algorithms such as SAC and TD3 were also unable to successfully learn to control the system. Our method, however, not only achieves exceptionally high returns from the outset but also continues to enhance performance within just a few hundred steps. Note that our curves only display the learning process within the first 1,000 steps, however for SAC and TD3, training extended beyond 80,000 steps, yet these algorithms consistently failed to acquire a successful control strategy for the IDP.

In addition, the results show that upon integration with our approach, every baseline improves sample efficiency and displays a reduced variance, indicating a more consistent learning process.


\subsubsection{Adaptability to Various Reward Functions}

The control module optimizes based on a presumed quadratic cost. However, as observed with the CartPole task, which uses a simple reward structure consisting only of a binary alive bonus (0 or 1), our method augments the performance of baselines, irrespective of the alignment between the environment's reward and the control module's cost. This indicates our method's adaptability in learning the optimal policy specific to the environment's reward structure.

\subsubsection{Comparisons with LQR}
In the CartPole task, we initialized our partial model with settings that led to an unstable system configuration. We subjected LQR to the same initial conditions for a fair comparison, as illustrated in Figure \ref{cartpole}. Under these settings, LQR could only achieve a performance score of around 100, significantly lower than our method. In the IDP task, we benchmarked our learned policy against LQR with accurately defined model parameters. While LQR is generally capable of stabilizing the system, it struggles to maintain control under certain challenging initial conditions, failing to keep the system within the environmental constraints. Leveraging the adaptive capabilities inherent in RL, our method demonstrates the ability to learn and perform under these difficult conditions. We present the accumulated costs encountered in such scenarios in \cref{tab:performance_comparison}, where the vectors in the first column include the cart's linear velocity and the angular velocities of the first and second poles.

\subsection{Real-world experiments}

\begin{figure}[tbp]
    \centering
    \includegraphics[width=2.5in]{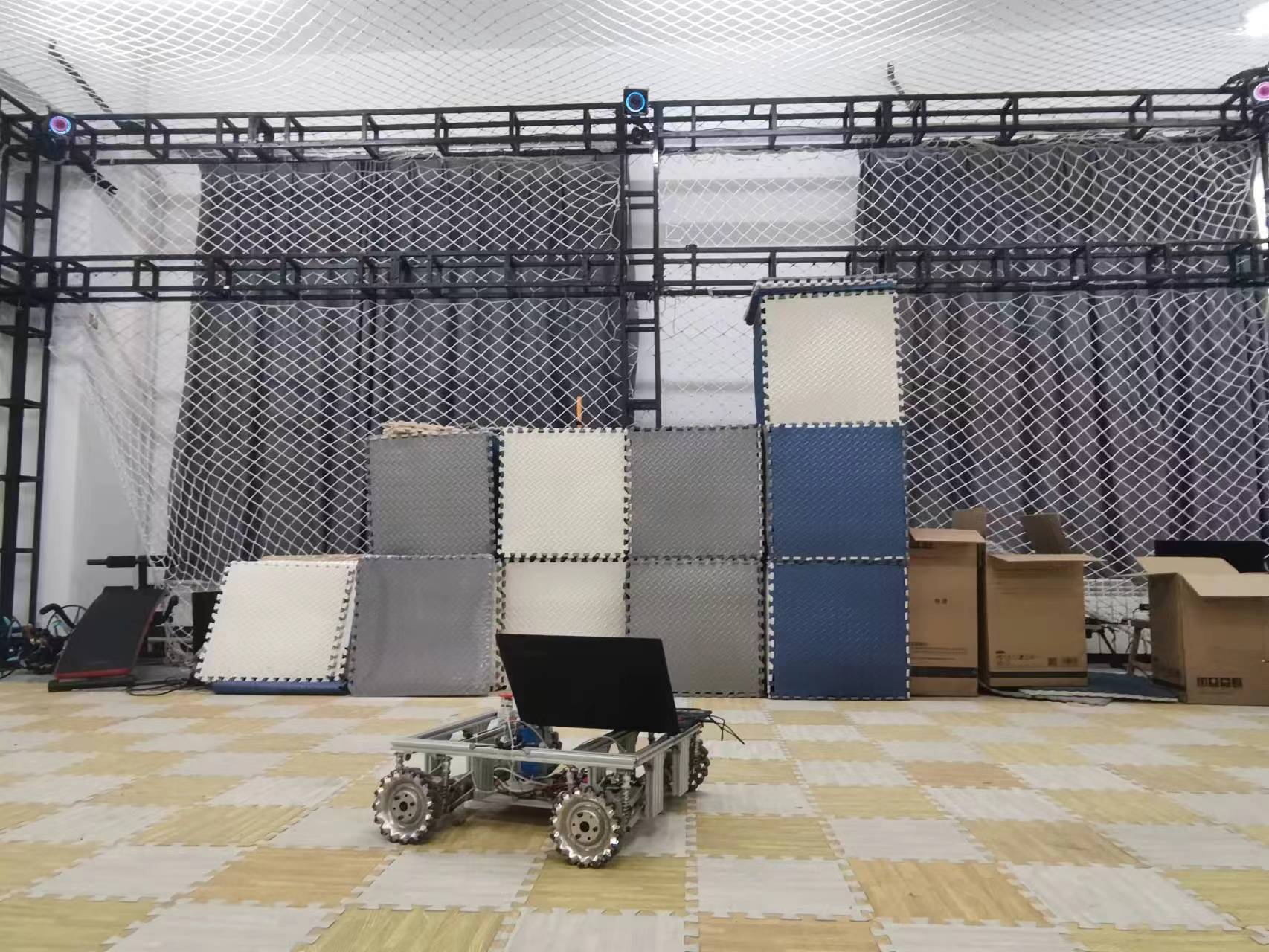}
    \caption{Experiment overview. The field is $6 \times 8\,$m, with $8$ cameras on top of the field capturing the pose and position of the robot. A target for the capturing system is fixed on the side of the robot.}
    \label{fig:robot}
    \vspace{-10pt}
\end{figure}

\begin{figure}[htbp!]
    \centering
    \begin{subfigure}[b]{1.6in}
        \centering
        \includegraphics[width=\textwidth]{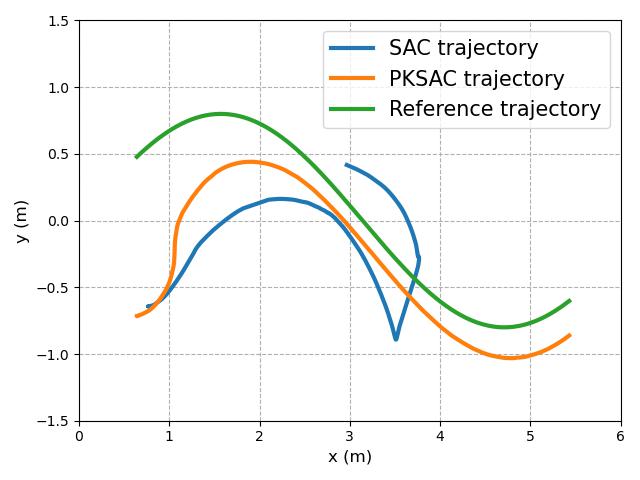}
        \caption{Trajectories starting below the sine wave}
        \label{traj1}
    \end{subfigure}
    \begin{subfigure}[b]{1.6in}
        \centering
        \includegraphics[width=\textwidth]{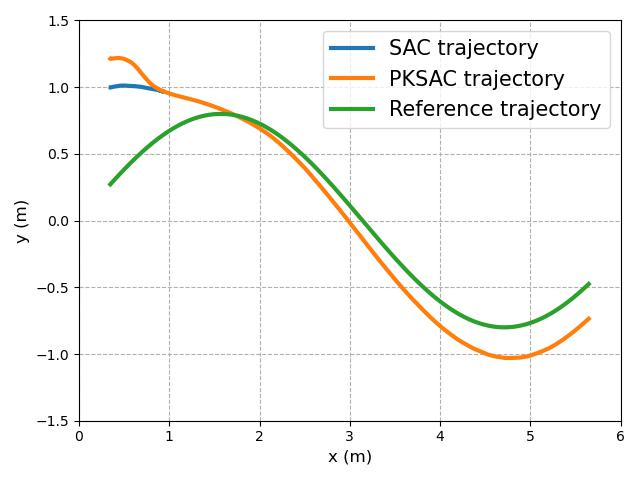}
        \caption{Trajectories starting above the sine wave}
        \label{traj2}
    \end{subfigure}
    \caption{Trajectories following our method, SAC, and the reference.}
    \label{traj}
    \vspace{-10pt}
\end{figure}
In our real-world experiments, we used a four-wheeled Mecanum robot to follow a sinusoidal path, ${y=0.8\sin(x)}$, aiming to achieve a consistent travel speed of $0.35\,$m/s along the desired path, from various initial positions far from the reference trajectory. We conducted two sets of experiments: one with the robot starting above the sine wave, and the other starting below, testing its ability to converge to the target speed along the trajectory. The tracking performance was defined as a quadratic cost function including the squared sum of the orientation error, deviation in the $y$ direction, and the difference from the desired tracking speed.

We use the OptiTrack Motion Capture System to capture the robot's positioning and orientation. An onboard computer running ROS processed the real-time data. The sampling interval for the controller was $0.1\,$s. The floor is covered with a foam mat, bringing hard-to-model divergence between simulation and reality.


The robot's state variables included its position $x$ and $y$, yaw angle $\theta$, linear velocity $v$, and angular velocity $\omega$. The controller generated outputs in terms of $\delta v$ and d$\delta \omega$, which denote the incremental changes in linear and angular velocities within the sampling interval. In this experiment, we implemented our approach with SAC as the foundational framework and compared its performance directly against the baseline SAC. The policies were pre-trained offline and deployed to the real world without any fine-tuning.
\begin{table}[hbp]
    \centering
    \caption{Tracking error under different scenarios and different methods.}
    \resizebox{0.35\textwidth}{!}{
    \begin{tabular}{l|ccc}
        \toprule 
        Scenario & PKSAC & SAC & Improvement (over SAC)\\
        \midrule 
        Upper start & $\textbf{126.27}$ & $173.68$ &$27.3\%$\\
        Lower start & $\textbf{72.55}$ & $120.72$ &$39.9\%$\\
        \bottomrule
    \end{tabular}}
    \label{tracking_error}
    \vspace{-10pt}
\end{table} 
\begin{figure}[htbp!]
    \centering
    \begin{subfigure}[b]{1.6in}
        \centering
        \includegraphics[width=\textwidth]{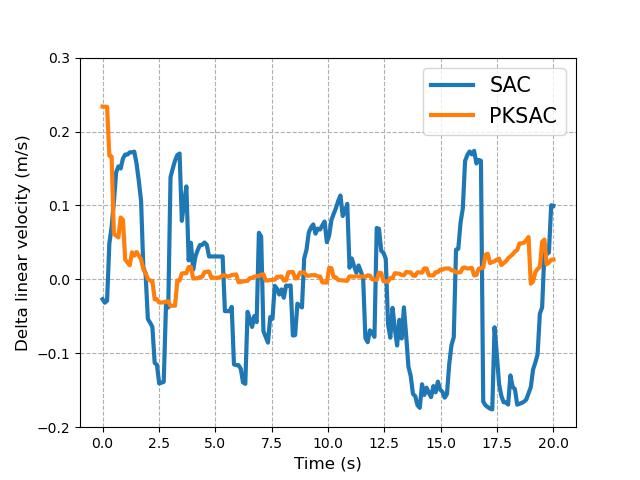}
    \end{subfigure}
    \begin{subfigure}[b]{1.6in}
        \centering
        \includegraphics[width=\textwidth]{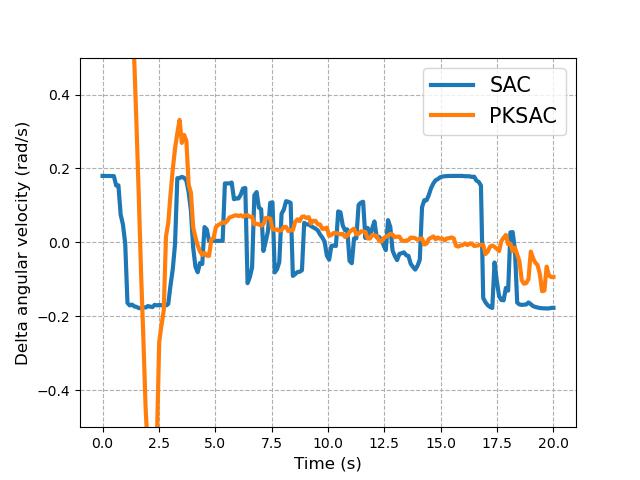}
    \end{subfigure}
    \caption{Actions $\delta v$ and $\delta \omega$ with time. Left is $\delta v$, right is $\delta \omega$. }
    \label{fig:delta}
\end{figure}
In \cref{traj}, it is clear that our method closely follows the desired sine curve trajectory, while the baseline SAC exhibits significant deviations due to real-world complexities such as wheel slippage and inaccuracies in lower-level controllers. Moreover, as illustrated in \cref{traj2}, SAC struggles with early termination; in certain start positions, it prematurely halts the trajectory to avoid accumulating further errors. This issue is observed in both simulations and real-world experiments. In contrast, our method demonstrates superior tracking precision. The incorporation of the control module enhances the robustness of our approach, effectively bridging the sim-to-real gap and guiding the robot along a higher-reward trajectory. Specifically, \cref{tracking_error} illustrates the reduction in accumulated tracking error achieved by our method.

A detailed examination of the robot's actions, as illustrated in \cref{fig:delta}, shows that our method produces actions with less fluctuation compared to SAC. This figure shows the actions for the case where the robot starts below the sine wave. Stable actions are essential for comfort and the health of the platform. This clear distinction accentuates the added value and efficacy of our control module.

In summary, the inherent challenges posed by the simulation-to-reality gap were more pronounced in the baseline SAC. In stark contrast, our extended method consistently demonstrated superior robustness and adaptability.

\section{Discussion}
One advantage of our method is that it doesn't break the framework of RL. Consequently, other methods dealing with safety and stability can be extended to our method, which is important for RL-based controller design. 
It is worth noting that if the system information is perfectly known, meaning that there is no model bias and all the parameters inside the dynamics are accurate, our framework reduces to a control framework.

Our method is currently restricted to regulation and tracking style problems and requires a differentiable model. Such tasks are often fit for an LQR controller. For more complex tasks such as the locomotion of HalfCheetah, and Humanoid that contain discontinuous contact force, our method currently has limitations.  We leave the adaption towards more complex tasks as future work.

\section{Conclusion}
In control tasks, some partial parametric model information is often known, but seemingly under-utilized in model-free RL.
We have developed a flexible framework that allows RL to leverage such information solely by adapting the policy network. Experiments show that our framework bridges RL and control theory, significantly improving sample efficiency and sim-to-real transfer compared to RL alone, and enhancing performance over conventional control methods. Future work will explore incorporating such information into the value network as well and extending our method to more complex tasks, such as locomotion and vision-based control.

\bibliographystyle{ieeetr}
\bibliography{reference.bib}






\end{document}